\documentclass[10pt,twocolumn,letterpaper]{article}

\usepackage{cvpr}              

\usepackage{graphicx}
\usepackage{amsmath}
\usepackage{amssymb}
\usepackage{booktabs}
\usepackage[dvipsnames]{xcolor}
\usepackage{pifont}
\usepackage{xspace}
\usepackage{multirow}
\usepackage{rotating}
\usepackage{color}
\usepackage[dvipsnames]{xcolor}
\usepackage{paralist}
\usepackage{tabularx, colortbl}
\usepackage[all=normal, mathspacing=tight, floats=tight, paragraphs=tight]{savetrees}

\usepackage[pagebackref,breaklinks,colorlinks]{hyperref}

\renewcommand{\b}{\boldsymbol}
\newcommand{\m}{\mathcal}
\DeclareMathOperator*{\argmax}{arg\,max}

\definecolor{ood}{rgb}{0.95,0.95,0.95}
\newcolumntype{a}{>{\columncolor{ood}}l}

\usepackage[capitalize]{cleveref}
\crefname{section}{Sec.}{Secs.}
\Crefname{section}{Section}{Sections}
\Crefname{table}{Table}{Tables}
\crefname{table}{Tab.}{Tabs.}

\def\cvprPaperID{9} 
\def\confName{UNCV-W}
\def\confYear{2022}

\begin{document}

\title{On the Usefulness of Deep Ensemble Diversity for Out-of-Distribution Detection}

\author{Guoxuan Xia  \\
Imperial College London \\
{\tt\small g.xia21@imperial.ac.uk}
\and
Christos-Savvas Bouganis \\
Imperial College London\\
{\tt\small christos-savvas.bouganis@imperial.ac.uk}
}
\maketitle

\begin{abstract}
The ability to detect Out-of-Distribution (OOD) data is important in safety-critical applications of deep learning. The aim is to separate In-Distribution (ID) data drawn from the training distribution from OOD data using a measure of uncertainty extracted from a deep neural network. Deep Ensembles are a well-established method of improving the quality of uncertainty estimates produced by deep neural networks, and have been shown to have superior OOD detection performance compared to single models. An existing intuition in the literature is that the \emph{diversity} of Deep Ensemble predictions indicates distributional shift, and so measures of diversity such as Mutual Information (MI) should be used for OOD detection. We show experimentally that this intuition is not valid on ImageNet-scale OOD detection -- using MI leads to 30-40\% worse \%FPR@95 compared to \emph{single-model} entropy on some OOD datasets. We suggest an alternative explanation for Deep Ensembles' better OOD detection performance -- OOD detection is binary classification and we are ensembling diverse classifiers. As such we show that practically, even better OOD detection performance can be achieved for Deep Ensembles by averaging task-specific detection scores such as Energy over the ensemble. We make our code available.\footnote{\url{https://github.com/Guoxoug/ens-div-ood-detect}}
\end{abstract}

\section{Introduction}\label{sec:intro}
Detecting Out-of-Distribution (OOD) data \cite{Yang2021GeneralizedOD}, i.e. input data that does not originate from the same distribution a deep neural network (DNN) is trained on, is a task that is attracting an increasing amount of attention in the domain of deep learning for computer vision \cite{Liang2018EnhancingTR,Liu2020EnergybasedOD,Du2022VOSLW,Hendrycks2017ABF,Hendrycks2019BenchmarkingNN,Fort2021ExploringTL,Hsu2020GeneralizedOD, Techapanurak_2020_ACCV,Sun2021ReActOD,Wang2022ViMOW,Huang2021MOSTS,Lee2018ASU,Pearce2021UnderstandingSC,Yang2021GeneralizedOD,Zhang2021OnTO,Nalisnick2019DoDG}. This is important for applications where passing predictions on OOD data downstream may incur a large cost, \eg in medicine or autonomous driving. 

Deep Ensembles \cite{Lakshminarayanan2017SimpleAS} have been shown to reliably be superior for OOD detection compared to single models \cite{Malinin2020EnsembleDD, Ryabinin2021ScalingED, Kim2021AUB}. An existing intuition for this is that the diversity of a Deep Ensemble is an intrinsic indicator of input distributional shift, as the members will disagree more on data that are further away from training data \cite{Malinin2020EnsembleDD, Mukhoti2021DeterministicNN}. In this work we: 
\begin{inparaenum}
    \item Show that this intuition is not valid for OOD detection for convolutional neural networks (CNNs) at ImageNet-scale.
    \item Present an alternative explanation for Deep Ensembles' superior performance -- they are ensembles of binary classifiers that are diverse in their OOD detection errors.
    \item Show that even better OOD detection performance can be achieved by averaging a task specific score such as Energy \cite{Liu2020EnergybasedOD} over a Deep Ensemble.
\end{inparaenum}
\section{Preliminaries}
\subsection{OOD Detection}
We use the problem setting defined in \cite{Yang2021GeneralizedOD} for OOD detection. For a $K$-class classification problem, we wish to predict labels $y\in\m Y=\{\omega_k\}_{k=1}^K$ given inputs $\b x\in \m X = \mathbb{R}^D$. We train a DNN with parameters $\b \theta$ on finite dataset $\m D_\text{train} = \{y^{(n)},\b x^{(n)}\}_{n=1}^{N}$ sampled independently from the true joint data distribution $p_\text{train}(y,\b x)$. The network typically models the conditional categorical distribution $P(y|\b x;\b\theta)$ using a softmax output layer, and is trained using cross entropy loss. Predictions $\hat y$ given test inputs $\b x^*$ can then be made as $\hat y = \argmax_\omega P(\omega|\b x^*;\b\theta)$.

During deployment, we expect $\b x^*$ to be sampled either from the training input distribution $p_\text{train}(\b x)$, or $p_\text{OOD}(\b x)$, where the \emph{label} space of the latter distribution has no intersection with $\m Y$. Predictions on the OOD samples will thus be invalid. In order to detect these samples the objective is then to design a binary classification function,
\begin{equation}\label{eq:bin}
    g(\b x;t) = \begin{cases}
    \text{ID}, &\text{if }U(\b x) < t\\
    \text{OOD}, &\text{if }U(\b x) \geq t~,
    \end{cases}
\end{equation}
where $U$ is a predictive uncertainty score (or $-U$ confidence), commonly extracted from the network. The performance can be evaluated using standard metrics for binary classification such as Area Under the ROC curve (AUROC$\uparrow$) and False Positive Rate at 95\% True Positive Rate (FPR@95$\downarrow$), where arrows ($\uparrow\downarrow$) indicate whether higher or lower is better. The problem is challenging, as the OOD data distribution $p_\text{OOD}(\b x)$ is not known to the practitioner before deployment.
\subsection{Deep Ensembles}
Deep Ensembles \cite{Lakshminarayanan2017SimpleAS} are ensembles of DNNs trained on the same dataset with different random parameter initialisations and data shuffling, $\Theta = \{\b \theta^{(m)}\}_{m=1}^M$. The ensemble members can be interpreted as samples from different modes of the Bayesian parameter posterior $p(\b \theta|\m D_\text{train})$ \cite{izmailov2021bayesian}, which allows their predictions to be more diverse compared to unimodal approximate Bayesian approaches \cite{Fort2019DeepEA}. They have been shown to be a reliable and scalable approach for improving both the predictive performance and quality of uncertainty estimates of DNNs \cite{Lakshminarayanan2017SimpleAS,ovadia2019can,Malinin2020EnsembleDD,Ryabinin2021ScalingED}.

Following the Bayesian interpretation of Deep Ensembles, we obtain an approximate posterior predictive distribution by averaging over the softmax outputs of the individual ensemble members,\footnote{We use a Bayesian framework as it provides convenient metrics, as such we are not particularly concerned with the fidelity of the approximation in \cref{eq:post}. Discussion of this can be found in \cite{wilson_2019, izmailov2021bayesian}.}
\begin{equation}\label{eq:post}
      P(y|\b{x},\mathcal{D}_{\text{train}}) \approx \frac{1}{M}\sum_{m=1}^M P(y|\b{x}, \b{\theta}^{(m)}) = P(y|\b{x}, \Theta),
\end{equation}
as well as an uncertainty decomposition \cite{Depeweg2018DecompositionOU} in the form of

\begin{multline}\label{eq:approx_decomp}
   \underbrace{\mathcal{I}[y, \b{\theta}|\b{x}, \mathcal{D}_{\text{train}}]}_{\text{mutual information}}  \approx  \underbrace{\mathcal{H}\left[\frac{1}{M}\sum_{m=1}^M P(y|\b{x}, \b{\theta}^{(m)})\right]}_{\text{ensemble entropy}}\\  -\underbrace{\frac{1}{M}\sum_{m=1}^M\mathcal{H}\left[P(y|\b{x}, \b{\theta}^{(m)})\right]}_{\text{average entropy}},
\end{multline}
where $\m H[\cdot]$ is the entropy of a probability distribution and $\mathcal{I}[\cdot,\cdot|\cdots]$ the (conditional) mutual information (MI) between two random variables. We will refer to the other terms in the above equation as ensemble entropy (Ens.$\m H$) and average entropy (Av.$\m H$). MI, when approximated using \cref{eq:approx_decomp}, measures the \emph{diversity} of the ensemble, as it is the mean KL-divergence between the predictive distributions of the ensemble members and the ensemble overall,
\begin{multline}\label{eq:kl}
    \mathcal{I}[y, \b{\theta}|\b{x}, \mathcal{D}_{\text{train}}] \approx \\
      \frac{1}{M}\sum_{m=1}^M\left[\text{KL}\left[P(y|\b{x}, \b{\theta}^{(m)})||P(y|\b{x},\Theta)\right]\right].
\end{multline}
Av.$\m H$ is a measure of \textit{average uncertainty} over the ensemble, and we can say that Ens.$\m H$ measures \textit{total uncertainty} by combining \textit{diversity} and \textit{average uncertainty}. Note in the following text, when we refer to MI, it is referring to the right hand side of \cref{eq:kl}, i.e. the empirical approximation.

Note that we will avoid the discussion of epistemic and aleatoric uncertainty \cite{Hllermeier2021AleatoricAE, Mukhoti2021DeterministicNN, Pearce2021UnderstandingSC, Malinin2020EnsembleDD} in this work and instead directly focus the discussion on the task of OOD detection.
\subsection{Experimental Setup}
In order to construct our Deep Ensemble, we train five ResNet-50 \cite{He2016DeepRL} models independently using random seeds $\{1,\dots,5\}$ and standard ImageNet settings.\footnote{We use the default settings found here \url{https://github.com/pytorch/examples/blob/main/imagenet/main.py}. For full details see \cref{app:training}} 

For our ID dataset we use ImageNet-200 \cite{Kim2021AUB}, which uses a subset of 200 ImageNet-1k \cite{Russakovsky2015ImageNetLS} classes. We use a large selection of ImageNet-scale OOD datasets that exhibit a wide range of semantics and difficulty in being identified. Near-ImageNet-200 \cite{Kim2021AUB} is composed of ImageNet-1k classes semantically similar but disjunct to ImageNet-200. Caltech-45 \cite{Kim2021AUB} is a subset of the Caltech-256 \cite{Griffin2007Caltech256OC} dataset with disjoint classes to ImageNet-200. Openimage-O \cite{Wang2022ViMOW} is a subset of the Open Images V3 \cite{openimages} dataset chosen to be OOD to ImageNet-1k. 
iNaturalist \cite{Huang2021MOSTS} and Textures \cite{Wang2022ViMOW} are the same for their respective original datasets \cite{inat, cimpoi14describing}. Colorectal \cite{Kather2016MulticlassTA} is composed of histological photographs of human colorectal cancer, whilst Colonoscopy is constructed from individual frames extracted from colonoscopic video of gastrointestinal lesions \cite{Mesejo2016ComputerAidedCO}. Noise is a collection of square images where the pixel intensities, image resolution and image contrast are randomly generated (for details see \cref{app:images}). Finally, ImageNet-O \cite{Hendrycks2021NaturalAE} is adversarially constructed by selecting OOD images that a trained ResNet makes confident predictions on. Example images from each dataset, as well as their sizes can be found in \cref{app:images}.

\section{Existing Intuition Doesn't Hold Up}
\begin{figure}
    \centering
    \includegraphics[width=.75\linewidth]{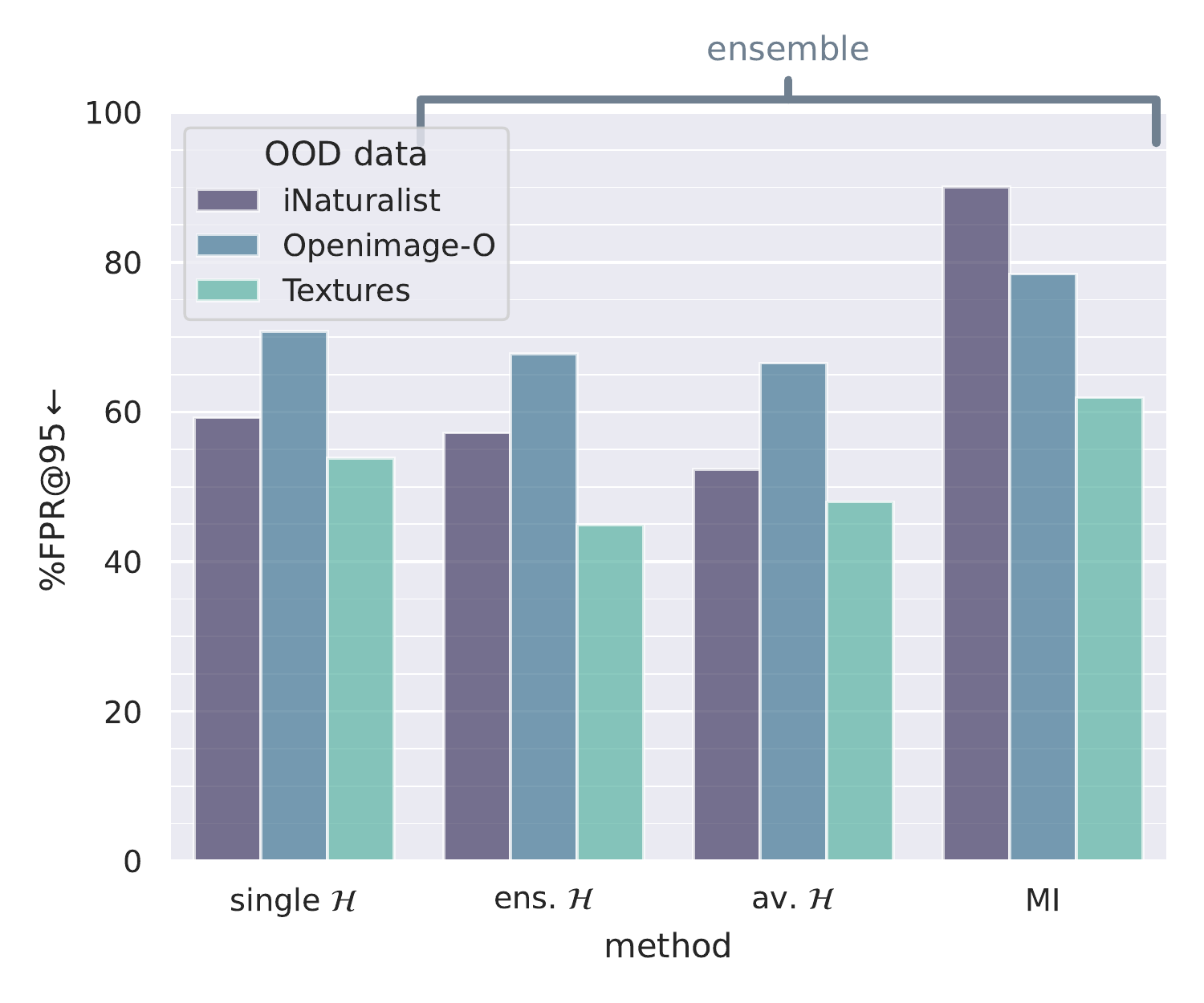}
    \caption{\%FPR@95$\downarrow$ for different uncertainty metrics. MI (measuring diversity) performs poorly across all OOD datasets. Ens.$\m H$ and Av.$\m H$ have similar performance and are able to improve over the single model baseline, although Av.$\m H$ is better in 2/3 cases. This suggests MI is not reliably useful by itself, or as part of Ens.$\m H$. The ID dataset is ImageNet-200, the model ResNet-50.}
    \label{fig:MI_illust}
\end{figure}

\begin{figure*}
    \centering
    \includegraphics[width=.6\textwidth]{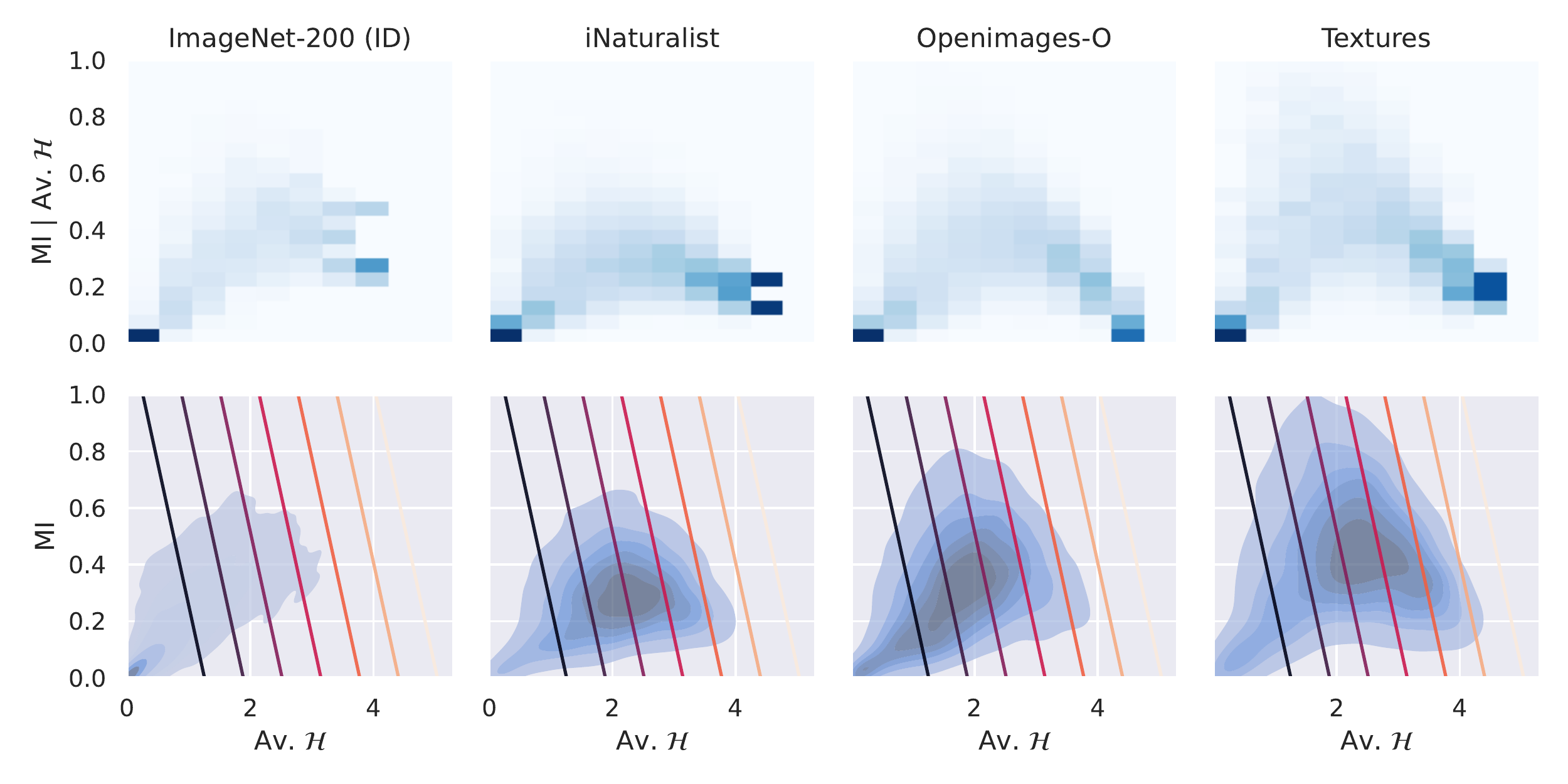}
    \caption{Top: empirical conditional histograms of MI$|$Av.$\m H$. The behaviour of MI is quite similar between ID and OOD data, with the exception of Textures being slightly higher. Bottom: joint kernel density estimate plots of MI and Av.$\m H$. Contours represent increasing (lighter) values of Ens.$\m H$. MI is low for high Av.$\m H$ OOD data. MI does not contribute significantly to Ens.$\m H$.}
    \label{fig:conditional}
\end{figure*}

In the literature there is an existing intuition, that the \emph{diversity} of a Deep Ensemble intrinsically indicates distributional shift of the input $\b x$. Thus, uncertainty scores $U$ directly derived from diversity should be useful for OOD detection \cite{Malinin2020EnsembleDD,Ryabinin2021ScalingED,Mukhoti2021DeterministicNN}. Moreover, consequently the superiority of measures of total uncertainty compared to single models (\eg Ens.$\m H$) should be due to their diversity component (MI in \cref{eq:approx_decomp}). The intuition relies on the idea that as the input data moves further away from the training distribution, the ensemble members will be less constrained by the training data, and so will behave differently to each other. This intuition tends to be demonstrated on toy, low-dimensional data \cite{Malinin2020EnsembleDD, Mukhoti2021DeterministicNN}. However, we will show that it doesn't hold for ImageNet-scale OOD detection with CNNs.

\cref{fig:MI_illust} shows \%FPR@95$\downarrow$ for different uncertainty measures $U$ over three OOD datasets (full results can be found in \cref{tab:results}). We see that although Ens.$\m H$ and Av.$\m H$ are both superior to the average performance of single-model $\m H$, MI is unable to outperform the single model baseline. In fact, on iNaturalist, it is $\sim$30\% worse. This shows that directly using the diversity of the ensemble to detect OOD data is actually a poor choice in this scenario. We note that inferior MI performance, at ImageNet-scale, has been shown before in \cite{Ryabinin2021ScalingED}. Additionally, the performance is very close between Av.$\m H$ and Ens.$\m H$, but Av.$\m H$ beats Ens.$\m H$ in 2/3 datasets, suggesting MI is a hindrance there. Similar behaviour over most of the other datasets can be observed in \cref{tab:results}.

\subsection{Mutual Information Performs Poorly}
In order to understand the results in \cref{fig:MI_illust} and \cref{tab:results}, we can consider empirical plots of the \emph{conditional} distribution of MI \textit{given} Av.$\m H$, as well as the joint distribution. These are shown for ImageNet-200 (ID) and various OOD datasets in \cref{fig:conditional}. The graded contours represent different values of Ens.$\m H$ (lighter means higher), as it is the sum of Av.$\m H$ and MI (\cref{eq:approx_decomp}). Av.$\m H$ is much lower on OOD data -- see \cite{Pearce2021UnderstandingSC} for a discussion of why. Intuitively, if the average uncertainty is high, then it is difficult for an ensemble to be diverse, as for high diversity the members need to disagree confidently (\cref{eq:approx_decomp}). Thus, the MI of high Av.$\m H$ OOD data is actually quite low, leading to MI's poor performance. In other words, MI is failing because Av.$\m H$ is performing well. 

The notable exception is ImageNet-O, where MI is able to perform the best. We hypothesise that this is due to the adversarial nature of the ImageNet-O data causing the members to be individually more confident (and thus diverse) on the OOD data. A similar effect is observed in \cite{Malinin2021UncertaintyEI} for neural machine translation where diversity measures are able to successfully detect pathological overconfidence when models copy OOD inputs rather than translating them.

\subsection{Mutual Information Does Not Reliably Contribute to Ensemble Entropy}

 We see that in \cref{fig:conditional} (top) for a \emph{given} value of Av.$\m H$, the behaviour of MI is not that different between the ID and OOD datasets -- the diversity does not seem to act as a strong indicator of distributional shift \textit{in addition} to the average uncertainty. It can also be observed that the actual variation in MI is much lower in comparison to Av.$\m H$. Thus, Ens.$\m H$ is in fact dominated by Av.$\m H$, which explains why their performance is so close. On Textures Ens.$\m H$ performs better compared to Av.$\m H$, and this suggests that MI is useful in addition to Av.$\m H$. \cref{fig:conditional} shows that MI given Av.$\m H$ is slightly higher on Textures. However, on the other two datasets the opposite is the case -- including diversity in addition to average uncertainty harms performance, as the conditional distributions are very similar between ID and OOD. 

We note similar results to \cref{fig:conditional} are shown in \cite{Abe2022DeepEW}, however the problem framing and conclusions reached in that work are different to ours.
\section{An Alternative Explanation}
\begin{table*}
    \centering
 \resizebox{.75\textwidth}{!}{   
\begin{tabular}{lllaallllllll}
\toprule
 & \textbf{ID:} & \textbf{ImageNet-200} & \multicolumn{2}{c}{\textbf{OOD mean}} & \multicolumn{2}{c}{\textbf{Near-ImageNet-200}} & \multicolumn{2}{c}{\textbf{Caltech-45}} & \multicolumn{2}{c}{\textbf{Openimage-O}} & \multicolumn{2}{c}{\textbf{iNaturalist}} \\
\textbf{Model} &  & \textbf{Method} & AUROC$\uparrow$ & FPR@95$\downarrow$ & AUROC$\uparrow$ & FPR@95$\downarrow$ & AUROC$\uparrow$ & FPR@95$\downarrow$ & AUROC$\uparrow$ & FPR@95$\downarrow$ & AUROC$\uparrow$ & FPR@95$\downarrow$ \\
\midrule
\multirow[c]{8}{*}{\begin{sideways}\shortstack[l]{\textbf{ResNet-50} \\ens. \%Err: 16.28 \\ av. \%Err: 19.01}\end{sideways}} & \multirow[c]{3}{*}{\begin{sideways}\textbf{Single}\end{sideways}} & MSP & 84.47 & 66.55 & 77.56 \scriptsize ±1.2 & 82.21 \scriptsize ±1.1 & 84.54 \scriptsize ±1.2 & 67.83 \scriptsize ±4.0 & 85.09 \scriptsize ±1.0 & 71.43 \scriptsize ±1.8 & 88.40 \scriptsize ±0.9 & 60.83 \scriptsize ±2.9 \\
 &  & $\mathcal{H}$ & 85.98 & 62.12 & 78.18 \scriptsize ±1.2 & 82.33 \scriptsize ±1.7 & 86.13 \scriptsize ±1.0 & 63.96 \scriptsize ±3.6 & 86.21 \scriptsize ±1.0 & 70.76 \scriptsize ±2.0 & 89.64 \scriptsize ±0.8 & 59.22 \scriptsize ±3.5 \\
 &  & Energy & \underline{88.56} & \underline{47.71} & 75.96 \scriptsize ±1.2 & 82.57 \scriptsize ±2.1 & \underline{88.23 \scriptsize ±0.8} & \underline{54.32 \scriptsize ±3.7} & 86.43 \scriptsize ±0.6 & \underline{62.24 \scriptsize ±1.6} & 90.78 \scriptsize ±0.4 & \underline{46.18 \scriptsize ±1.9} \\
 \cmidrule{2-13}
 & \multirow[c]{5}{*}{\begin{sideways}\textbf{Ens.}\end{sideways}} & MSP & 86.67 & 63.53 & \underline{79.28} & 81.27 & 86.71 & 64.48 & 87.24 & 68.39 & 89.95 & 59.15 \\
 &  &Ens. $\mathcal{H}$ & \underline{88.21} & 57.97 & \underline{79.70} & \underline{80.85} & 88.15 & 60.04 & \underline{88.14} & 67.75 & \underline{91.05} & 57.19 \\
 &  & Av. $\mathcal{H}$ & 88.07 & \underline{57.48} & \textbf{79.75} & \textbf{80.40} & \underline{88.22} & \underline{59.31} & \underline{88.29} & \underline{66.50} & \underline{91.67} & \underline{52.31} \\
 &  & MI & 83.46 & 78.84 & 77.94 & 84.18 & 83.75 & 79.36 & 83.46 & 78.40 & 81.97 & 90.05 \\
 &  & Av. Energy & \textbf{90.16} & \textbf{43.38} & 77.60 & \underline{80.47} & \textbf{89.99} & \textbf{48.23} & \textbf{88.30} & \textbf{57.44} & \textbf{92.43} & \textbf{40.14} \\
\bottomrule
\end{tabular}
}
\resizebox{.75\textwidth}{!}{
\begin{tabular}{lllllllllllll}
\toprule
& \textbf{ID:} & \textbf{ImageNet-200} & \multicolumn{2}{c}{\textbf{Textures}} & \multicolumn{2}{c}{\textbf{Colonoscopy}} & \multicolumn{2}{c}{\textbf{Colorectal}} & \multicolumn{2}{c}{\textbf{Noise}} & \multicolumn{2}{c}{\textbf{ImageNet-O}} \\
\textbf{Model} &  & \textbf{Method} & AUROC$\uparrow$ & FPR@95$\downarrow$ & AUROC$\uparrow$ & FPR@95$\downarrow$ & AUROC$\uparrow$ & FPR@95$\downarrow$ & AUROC$\uparrow$ & FPR@95$\downarrow$ & AUROC$\uparrow$ & FPR@95$\downarrow$ \\
\midrule
\multirow[c]{8}{*}{\begin{sideways}\shortstack[l]{\textbf{ResNet-50} \\ens. \%Err: 16.28 \\ av. \%Err: 19.01}\end{sideways}} & \multirow[c]{3}{*}{\begin{sideways}\textbf{Single}\end{sideways}} & MSP & 86.95 \scriptsize ±1.4 & 60.49 \scriptsize ±4.0 & 90.67 \scriptsize ±1.8 & 53.16 \scriptsize ±9.0 & 89.82 \scriptsize ±5.2 & 50.88 \scriptsize ±11.9 & 81.70 \scriptsize ±6.5 & 71.13 \scriptsize ±15.6 & 75.50 \scriptsize ±0.6 & 80.96 \scriptsize ±0.7 \\
 &  & $\mathcal{H}$ & 88.69 \scriptsize ±1.3 & 53.84 \scriptsize ±3.8 & 92.94 \scriptsize ±1.6 & 42.88 \scriptsize ±9.1 & 91.95 \scriptsize ±4.9 & 41.10 \scriptsize ±15.7 & 83.61 \scriptsize ±5.3 & \underline{64.78 \scriptsize ±22.8} & 76.51 \scriptsize ±0.7 & 80.18 \scriptsize ±0.3 \\
 &  & Energy & \underline{92.93 \scriptsize ±1.0} & \underline{32.65 \scriptsize ±2.8} & \underline{95.40 \scriptsize ±1.5} & \underline{28.66 \scriptsize ±12.1} & \underline{98.25 \scriptsize ±1.7} & \underline{9.87 \scriptsize ±9.7} & \underline{91.99 \scriptsize ±7.6} & \textbf{38.43 \scriptsize ±23.3} & 77.10 \scriptsize ±0.9 & \underline{74.47 \scriptsize ±1.3} \\
 
  \cmidrule{2-13}
 
 & \multirow[c]{5}{*}{\begin{sideways}\textbf{Ens.}\end{sideways}} & MSP & 90.19 & 53.49 & 92.93 & 46.49 & 90.98 & 49.22 & 84.43 & 71.29 & 78.30 & 77.95 \\
 &  &Ens. $\mathcal{H}$ & \underline{91.86} & \underline{44.92} & 95.22 & \underline{31.60} & 93.63 & 36.18 & \underline{86.95} & 65.79 & \underline{79.17} & 77.45 \\
 &  & Av. $\mathcal{H}$ & 91.23 & 48.00 & \underline{95.26} & 31.65 & \underline{93.97} & \underline{34.46} & 85.67 & 65.75 & 78.56 & 78.95 \\
 &  & MI & 88.43 & 61.90 & 89.11 & 76.07 & 82.14 & 88.66 & 84.92 & 77.39 & \textbf{79.41} & \underline{73.55} \\
 &  & Av. Energy & \textbf{94.55} & \textbf{27.07} & \textbf{96.87} & \textbf{19.31} & \textbf{99.08} & \textbf{4.48} & \textbf{94.00} & \underline{41.13} & \underline{78.64} & \textbf{72.15} \\
\bottomrule
\end{tabular}
}
    \caption{\%AUROC$\uparrow$ and \%FPR@95$\downarrow$ for different uncertainty scores over all OOD datasets. We show mean$\pm2$std. for single model results. \textbf{Bold} is best performance, \underline{underline} 2nd or 3rd best. MI performs poorly compared to other uncertainties and even single models. Ens.$\m H$ and Av.$\m H$ consistently improve over single model $\m H$. Av. Energy is able to improve over single model Energy and is best overall.}
    \label{tab:results}
\end{table*}
Uncertainty scores derived directly from ensemble diversity don't seem to reliably contribute to the OOD detection performance of Deep Ensembles at ImageNet-scale. However, the question still remains: \textit{why do Deep Ensembles perform better than single models for OOD detection?}
\subsection{Deep Ensembles as Binary Classifiers}
The fact that Av.$\m H$ of the ensemble is able to consistently improve OOD detection performance compared to single model $\m H$ (\cref{tab:results}) points to a straightforward explanation. If we consider again the binary classification model presented in \cref{eq:bin}, we can simply view Deep Ensembles for OOD detection through the same lens we view them for ID prediction. By being diverse in the individual errors they make for OOD detection, their overall OOD detection performance after combination will be better, and this diversity arises from the multimodal nature of the ensemble in parameter space \cite{Fort2019DeepEA}. For a more formal explanation we can consider constructing a binary probabilistic classifier arbitrarily from uncertainty score $U$,
\begin{equation}
    P(s=\text{OOD}|\b x, \b \theta)  = 1-P(s=\text{ID}|\b x,\b\theta) = aU(\b x;\b \theta) + b,
\end{equation}
where $s\in\{\text{ID},\text{OOD}\}$ and $a, b$ are arbitrary parameters such that $U$ is mapped to a valid probability. Note that averaging $U$ over the ensemble is equivalent to averaging $P(s|\b x;\b \theta)$.
Jensen's inequality for the binary classification negative log-likelihood (NLL) loss for OOD detection thus states,
\begin{equation}
    -\log \left(\frac{1}{M}\sum_{m=1}^M P(s|\b{x}, \b{\theta}^{(m)})\right) \leq - \frac{1}{M}\sum_{m=1}^M \log P(s|\b{x}, \b{\theta}^{(m)}).
\end{equation}
 We can thus see how averaging $U (= \m H)$ over the ensemble leads to better OOD detection as the NLL of the ensemble OOD detector will be lower than the average NLL of the individual ensemble members. 

The \emph{key intuition} is that by ensembling binary classifiers that are diverse such that they make OOD detection errors on different data samples, the ensemble members are able to compensate for each other's miss-predictions and raise the ensemble performance above their individual average.
\subsection{Using a Better OOD Detection Score}
Existing work benchmarking Deep Ensembles focuses exclusively on uncertainty scores $U$ extracted from the softmax outputs of the members. Given the explanation in the previous section, it is natural to ask whether we can improve performance by ensembling a better $U$. Using Energy \cite{Liu2020EnergybasedOD} ($U = -\log\sum_k\exp v_k$, where $\b v$ are the logits), and averaging over the ensemble, it can be seen in \cref{tab:results} that average ensemble Energy is able to improve over single model Energy, as well as produce the best OOD detection results in the whole comparison. We remark that although we have chosen Energy here, we expect any $U$ with better performance than single model $\m H$ to receive a similar uplift upon ensembling.
\section{Concluding Remarks}
In this work, we show that the existing intuition, that the diversity of a Deep Ensemble is a useful indicator for whether data is OOD, does not hold for ImageNet-scale CNNs. We instead suggest an alternative explanation for Deep Ensembles' superior OOD detection performance compared to single models: OOD detection is binary classification, and individual Deep Ensemble members produce diverse errors on this binary classification task, so combining them together yields better performance.

{\small
\bibliographystyle{ieee_fullname}
\bibliography{bib}
}
\clearpage
\appendix
\section{Dataset Information}\label{app:images}
\cref{fig:img} shows randomly sampled images from each dataset, as well as the number of samples in each.
Below we describe the methodology for constructing the Colonoscopy and Noise OOD datasets. We refer to the original papers of the remaining OOD datasets for their corresponding details \cite{Huang2021MOSTS, Wang2022ViMOW, Kim2021AUB, Hendrycks2021NaturalAE, Kather2016MulticlassTA}. We note, for ImageNet-200, there is a slight discrepancy between the number of samples reported in \cite{Kim2021AUB} and the number in the authors' provided datasets.\footnote{\url{https://github.com/daintlab/unknown-detection-benchmarks}} However, we do not believe this affects the validity of our results.
\subsection{Noise}
\begin{figure*}
    \centering
    \includegraphics[width=.7\linewidth]{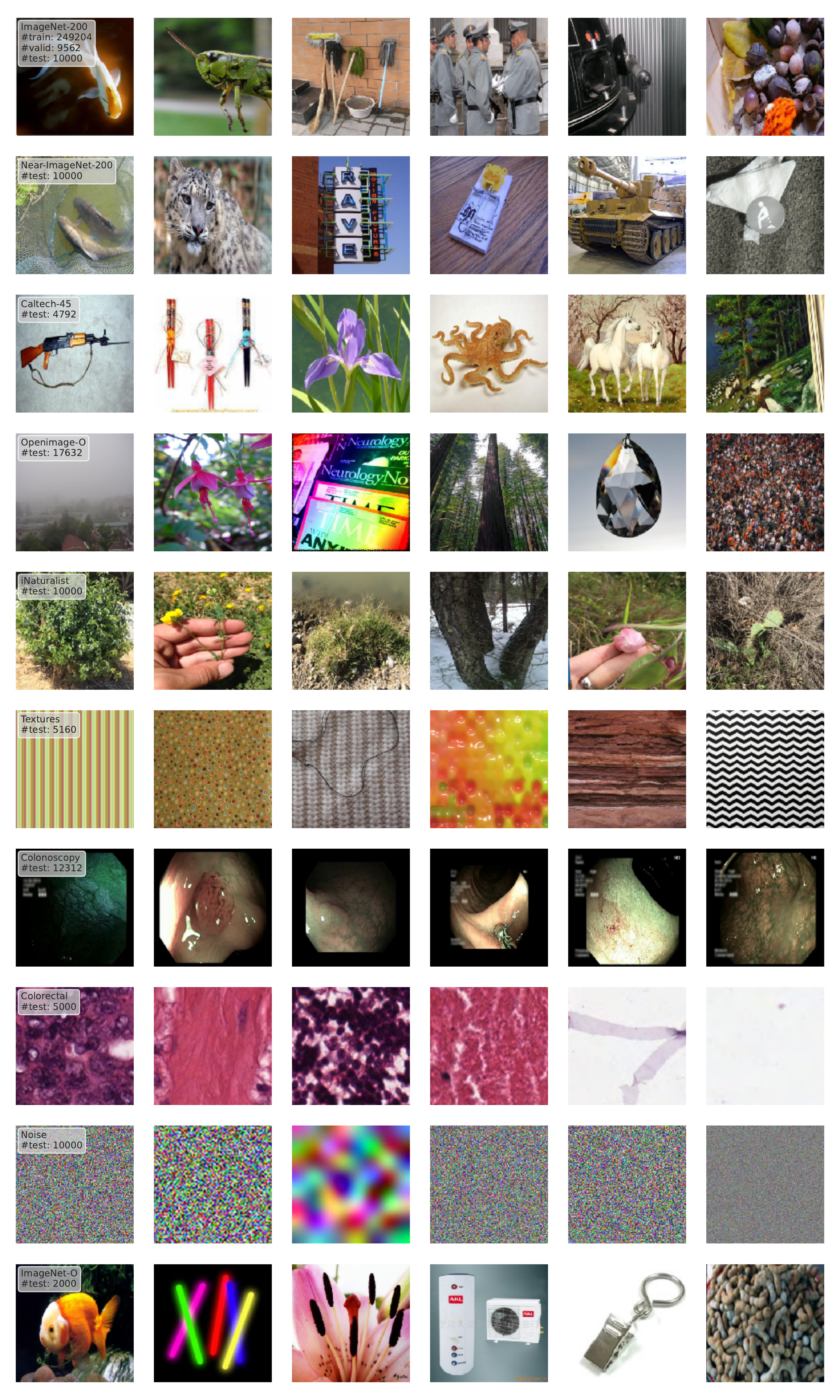}
    \caption{Example images from each dataset, with \#samples in each.}
    \label{fig:img}
\end{figure*}

10,000 square images are generated to form this dataset. All random samples are generated independently. For each image, each pixel intensity value (across x,y,R,G,B) is sampled from the same distribution.  It is a gaussian, $\m N(0.5, \sigma^2)$ where $\sigma$ differs \textit{between} images. $\sigma$ is generated by sampling from a unit gaussian and squaring the samples. Intensities are clipped to be in the range $[0,1]$ and mapped to integer values $\in \{0,...,255\}$. The image widths/heights are sampled from $\{2,...,256\}$ uniformly. Then, images are scaled to a resolution of $256\times256$ using the lanczos method.\footnote{\url{https://pillow.readthedocs.io/en/stable/_modules/PIL/Image.html\#Image.resize}} The resulting images therefore vary in contrast, intensity and scale (see \cref{fig:img}).
\subsection{Colonoscopy}
We separate out frames from the colonoscopical videos provided in \cite{Mesejo2016ComputerAidedCO}.\footnote{\url{http://www.depeca.uah.es/colonoscopy_dataset/}} The first 10 narrow band imaging (NBI) videos in each class of lesion (hyperplasic, serrated, adenoma) are downloaded. Each video frame is then extracted as an separate image to construct the dataset.

\section{Model Training}\label{app:training}
To construct the Deep Ensemble, 5 independent training runs are performed, with random seeds $\{1,...,5\}$.
ResNet-50 \cite{He2016DeepRL} is trained with the default hyperparameters in PyTorch's ImageNet example.\footnote{\url{https://github.com/pytorch/examples/tree/main/imagenet}} The ensemble members are trained on  ImageNet-200 for 90 epochs with batch size equal to 256. Stochastic gradient descent is used with an initial learning rate of 0.1 that steps down by a factor of 10 at epochs 30 and 60. Weight decay is set to $10^{-4}$, and momentum 0.9. During training images are augmented using \texttt{RandomHorizontalFlip} and \texttt{RandomResizedCrop}, and input to the network at a resolution of $224\times 224$. 

\end{document}